%% file: main.tex
\pdfoutput=1
\documentclass[letterpaper, 10 pt, conference]{ieeeconf}

\IEEEoverridecommandlockouts %
\usepackage{times}

\usepackage{multicol}
\usepackage{hyperref}
\usepackage{booktabs}
\usepackage{graphicx}
\usepackage{float}
\usepackage{amsmath, amssymb}
\usepackage{amsfonts}
\usepackage{bm}
\usepackage{algpseudocode}
\usepackage{algorithm}
\usepackage{multirow}
\usepackage{overpic}
\usepackage{tikz}

\usepackage[normalem]{ulem}
\usepackage[noadjust]{cite}

\newcommand\Reals{\mathbb{R}}

\renewcommand{\vec}{\mathbf}
\newcommand{\pref}{\vec{p}_\mathrm{ref}}

\usepackage{amsmath}

\DeclareMathOperator*{\argmin}{arg\,min}

\usepackage{color}

\newcommand{\annotatedFigure}[3][]{%
  \begin{tikzpicture}[above right, inner sep=0, outer sep=0]
    \node (image) at (0, 0) { \includegraphics[#1]{#3} };
    \node (label) at (1mm, 1mm) {#2};
  \end{tikzpicture}%
}

\title{Toward Learning Context-Dependent Tasks from Demonstration for Tendon-Driven Surgical Robots}

\input{author}

\begin{document}

\maketitle

\begin{abstract}
Tendon-driven robots, a type of continuum robot, have the potential to reduce the invasiveness of surgery by enabling access to difficult-to-reach anatomical targets.
In the future, the automation of surgical tasks for these robots may help reduce surgeon strain in the face of a rapidly growing population.
However, directly encoding surgical tasks and their associated context for these robots is infeasible.
In this work we take steps toward a system that is able to learn to successfully perform context-dependent surgical tasks by learning directly from a set of expert demonstrations.
We present three models trained on the demonstrations conditioned on a vector encoding the context of the demonstration.
We then use these models to plan and execute motions for the tendon-driven robot similar to the demonstrations for novel context not seen in the training set.
We demonstrate the efficacy of our method on three surgery-inspired tasks.

\end{abstract}

\section{introduction}
\input{introduction}

\section{Related work}\label{related-work}

\input{related-work}

\section{Problem Definition}

\input{problem-definition}

\section{Approach}\label{approach}
\input{approach}

\section{Experimental Results}\label{experiments}
\input{experiments}

\section{Discussion}\label{discussion}
\input{Discussion}

\section*{ACKNOWLEDGMENT}
The authors thank the group of D. Caleb Rucker for assistance with the tendon-driven robot kinematics, Dr. Chakravarthy Reddy for clinical insights, and Rahul Benny for assistance with visualization. This work was partially supported by NSF Awards \#2024778 and \#2133027. 

\bibliographystyle{IEEEtran}
\bibliography{references}

\end{document}

%% file: author.tex
\author{%
  Yixuan Huang,
  Michael Bentley,
  Tucker Hermans,
  and Alan Kuntz%
  \thanks{
    The authors are with the Robotics Center and School of Computing,
    University of Utah,
    Salt Lake City, UT 84112, USA. TH is also affiliated with NVIDIA.
    \protect\url{yixuan.huang@utah.edu},
    \protect\url{mbentley@cs.utah.edu},
    \protect\url{tucker.hermans@utah.edu}, and
    \protect\url{alan.kuntz@utah.edu}%
  }%
}

%% file: introduction.tex
Continuum robots, capable of taking curvilinear shapes, are a promising paradigm in minimally invasive surgery~\cite{Burgner2015_TRO}.
These robots are generally small in form and can curve around anatomical structures in the body enabling expressive motion and the ability to work in complex anatomy where surgical sites would be difficult to access with traditional straight instruments~\cite{Burgner2015_TRO,Webster2010_IJRR}.
Tendon-driven continuum robots are generally constructed of a long flexible backbone with tendons routed down their length through disks affixed to the backbone.
These tendons are then robotically actuated at the robot's base to change the shape of the backbone~\cite{Nguyen2015_IROS,Kato2015_TMECH,Kutzer2011_ICRA}.
Typically these tendons are routed straight down the backbone, producing constant curvature segments when pulled, however more complex tendon routing enables more complex shapes and expressive motion from the robot~\cite{Rucker2011_TRO,Oliver-Butler2019_TRO}.
With such non-linear tendon routing, the complex actuation of these robots makes automating their motion difficult.

\begin{figure}[t]
    \centering
    \includegraphics[width=\columnwidth]{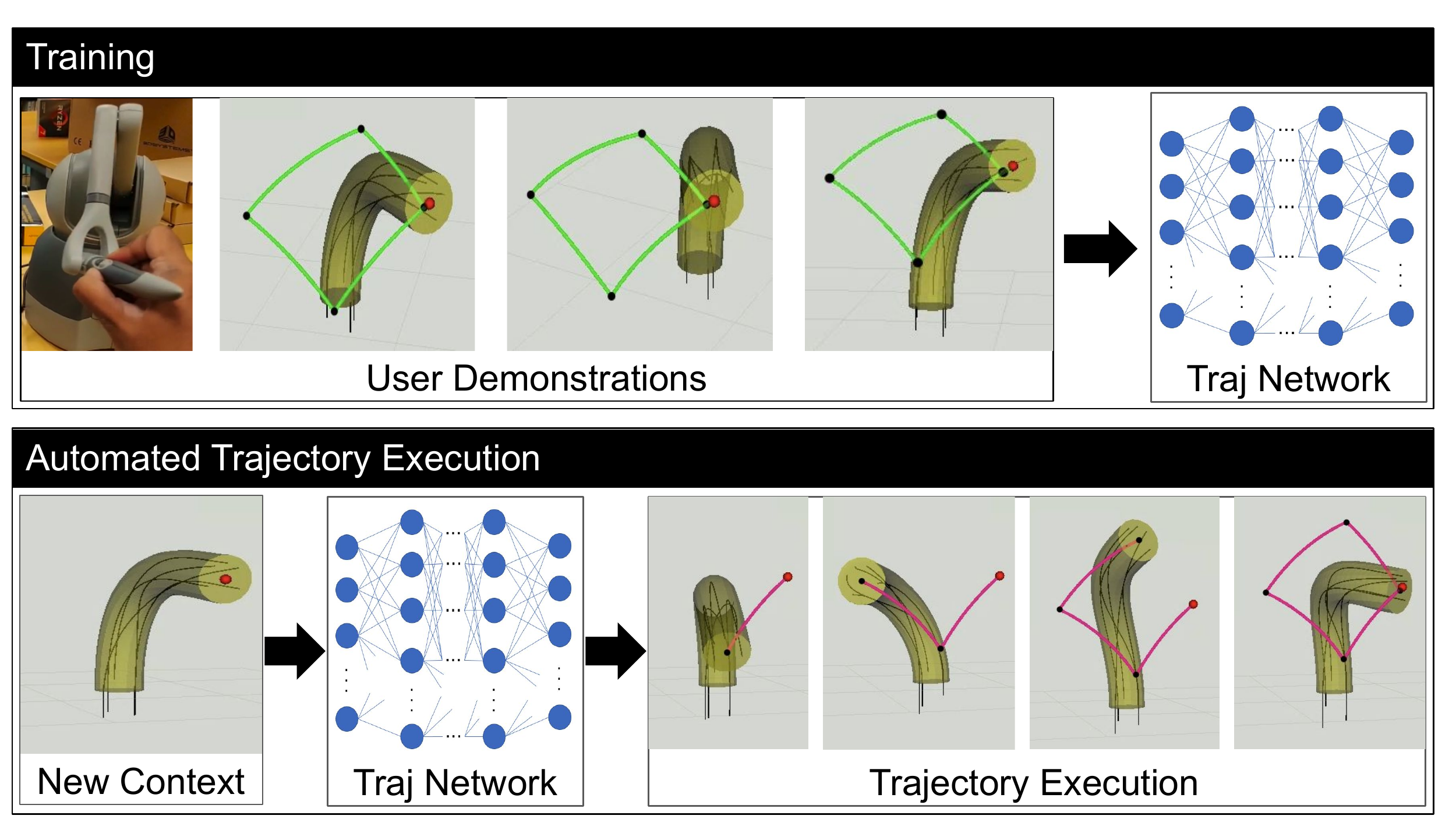}
    \vspace{-1.5em}
    \caption{Our method learns to perform tasks for a tendon-driven continuum robot from human demonstrations. Using a haptic input device, a human controls the robot to perform a task, such as moving the tip along a rectangular space curve (green). This is done multiple times with different contexts. In our approach, we use these demonstrations to train a trajectory network which enables the robot to perform the task for a novel context, such as in a new location or with differing scale~(pink).}
    \label{fig:high_level_figure}
    \vspace{-1.5em}
\end{figure}

Automating surgical-tasks is particularly challenging, even for more traditional robot types, due to the complex nature of encoding the tasks and their constraints into the automation system~\cite{yip2019robot}.
Further, the way in which the task should be executed varies based on the context, e.g., features of the patient's specific anatomy.
In this work, we take significant steps toward addressing these challenges by presenting a method that learns to perform tasks, dependent on context, from a set of human-performed expert demonstrations.
We collect expert demonstrations of task execution using a teleoperated complex-routed tendon-driven robot, in scenarios where context variables differ.
Our method then learns how the task execution varied with respect to the context from these demonstrations and is then capable of performing the task in environments where the context differs from the scenarios seen during training, e.g., when task relevant features are in new locations, task-relevant geometry has changed, etc.

To do so, we leverage aspects of Learning from Demonstration (LfD)~\cite{argall2009survey,ravichandar2020recent} and contextual learning~\cite{kober2011reinforcement, kumar2019contextual}.
This enables our method to learn to perform tasks from surgeon demonstrations without explicitly encoding the desired robot motion or encoding how it should change based on the context.
We collect a small set of 
expert demonstrations using a teleoperation system, in which a haptic device is used to control a tendon-driven robot in simulated medical scenarios as the task is completed by the expert user, where context is varied between demonstrations.
These demonstrations are then used to train a learned model that takes as input context variables and outputs a work-space trajectory, i.e., an ordered sequence of tip motions for the robot to perform, which is then executed on the robot via iterative inverse kinematic control.
We learn, evaluate, and compare three different models within our framework: (i) a linear approach, (ii) a non-linear kernel-based approach, and (iii) a feed-forward neural network.
Our approach, once trained on the expert demonstrations, is capable of producing trajectories that successfully perform the demonstrated task in novel situations (contexts) not seen during training (see Fig.~\ref{fig:high_level_figure}).

We demonstrate the efficacy of our method on three surgery-inspired tasks.
In the first two, we show the ability to trace expert-demonstrated curves, via a sequence of points---first on the surface of a plane and second on the surface of spheres---as proxy tasks for learning to cut the surface of tissue in specific ways with, e.g., electrocautery or laser ablation.
As context we vary the start-location of the desired curves as well as geometric aspects of the environment, e.g., the width/height of the plane and the radii of the spheres. 
In both cases, the curve the robot is tracing and the geometric constraints (i.e., staying on the surface of the sphere or plane), as encoded by the expert-generated robot motion, are learned entirely from the demonstrations and the context.
Our third task is inspired by pleuroscopy, a procedure in which a clinician is operating an endoscope between the chest wall and a collapsed lung~\cite{Michaud2010_Chest}.
Our method learns from demonstrations to navigate inside the pleural space, segmented from a real patient, and successfully trace a curve on the surface of the anatomy even when changing the position and scale of the anatomy with respect to the robot.

We evaluate the performance of our method with the three learned model approaches, varying parameters such as the neural network architecture and the number of expert demonstrations.
Notably, the neural network approach demonstrates performance close to human-level in some cases with relatively few demonstrations.

In this work, we take significant steps toward the automation of context-dependent surgical tasks learned from demonstration, removing the need to directly encode the tasks and their constraints.
This work represents the first use of contextual learning for producing complex trajectories for surgical robots and the first instance of LfD in continuum robots.
We show that with a relatively small number of demonstrations our learning-based method is capable of inferring the context-dependent task and its constraints solely from the demonstrations.
Once trained, our method is then able to perform the learned task, adhering to the learned constraints, in novel context/situations not seen during the demonstrations.

%% file: related-work.tex
Continuum robots have been proposed for a variety of medical tasks~\cite{Burgner2015_TRO}.
Tendon-driven robots are one example of a continuum robot with promising potential medical applications~\cite{Nguyen2015_IROS,Kato2015_TMECH,Kutzer2011_ICRA}.
Complex routing of the tendons enable these robots to take a variety of complex shapes as the tendons are actuated as well as achieve desired stiffness properties~\cite{Rucker2011_TRO,Oliver-Butler2019_TRO}.
In this work, we consider a tendon-driven robot with straight and helically routed tendons and utilize a version of the state-of-the-art kinematic model presented in~\cite{Rucker2011_TRO}, chosen for its ability to model the robot's shape given complex tendon routings.

Learning from Demonstration (LfD) is a branch of machine learning in which autonomous task execution is learned from task-specific human demonstrations~\cite{argall2009survey,ravichandar2020recent}.
LfD has shown impressive performance on non-medical robots and tasks including manipulation~\cite{conkey2019learning, akgun2012keyframe}, autonomous driving~\cite{silver2013learning, kuderer2015learning}, and bipedal robot locomotion~\cite{mericcli2010biped}.
Compared to traditional planning and control methods, LfD directly learns to successfully perform tasks from human demonstrations without the need to explicitly encode task specifics or constraints.
Thus LfD is a particularly promising paradigm for tasks that are difficult to encode.

In the medical domain, LfD has been applied in a variety of ways.
For instance, van den Berg et al. present an apprenticeship learning approach to solve a two-handed knot tying task~\cite{van2010superhuman}.
Murali et al. propose a learning-by-observation approach for autonomous multilateral medical tasks~\cite{murali2015learning}.
Kim et al. propose to leverage LfD to achieve automation of tool-navigation tasks in retinal surgery~\cite{kim2020autonomously}.
LfD has also been applied to provide therapy for patients~\cite{fong2018kinesthetic} and to assist children with cerebral impairments~\cite{najafi2017robotic}.
However, LfD has not yet been applied to the automation of surgical tasks for continuum robots, which is the subject of this work.

Machine learning methods have been applied to solving other problems for continuum robots.
For instance, data-driven approaches have been applied to learn the inverse kinematics of tendon-drive robots~\cite{xu2017data, giorelli2015neural}.
Data-driven methods have also been applied to concentric tube robots, used in learning the forward and inverse kinematics~\cite{Bergeles2015_Hamlyn,Grassmann2018_IROS}, the complete shape~\cite{kuntz2020learning}, and in estimating tip-contact forces~\cite{Donat2020_TMRB}.
Further, Iyengar et al. \cite{iyengar2020investigating} leverage deep reinforcement learning to control concentric tube robots, a method distinct from LfD.

Contextual learning has been applied to a variety of robotics tasks in other domains. 
For instance, Kumar et al.~\cite{kumar2019contextual} formulate a multi-finger grasping task as a contextual policy search problem.  
Kober et al.~\cite{kober2011reinforcement} propose to generalize elementary movements by changing the meta-parameters of primitives in a context learning framework. 

In this work, we build upon existing LfD and contextual learning methods from other robotics domains.
This enables our method to learn context-dependent surgical tasks for tendon-driven robots.

%% file: problem-definition.tex
We consider a tendon-driven robot with $N$ tendons that travel down the length of the robot with arbitrary routing.
Each tendon can be pulled at the robot's base changing the tendon's tension and affecting the shape of the robot according to its routing, with the tension of tendon $i$ defined as $\tau_i \in [0,\tau_i^{\mathrm{max}}]$, a maximum tension value.
The robot can also be inserted and retracted, with its insertion length defined as $\ell \in [0, \ell^{\mathrm{max}}]$, the maximum insertion length of the robot.
The robot is also capable of being rotated at its base, with its rotation defined as $\beta \in [-\pi, \pi)$. %

A configuration for the robot then is defined as the vector $\vec{q} = [(\tau_i: i = 1,\ldots,N), \ell, \beta]$ with configuration space $\mathcal{Q} = \mathbb{R}_+^{N+1} \times S^1$.
A configuration can then be mapped to the robot's shape, including its tip pose, using the forward kinematics (FK) function, and tip pose mapped to a configuration via inverse kinematics (IK).

We next define a workspace trajectory consisting of $M$ 3D waypoints for the robot's tip as an ordered sequence $T = \langle p_1, p_2,\dots, p_{M}\rangle,$ $ p \in \Reals^3.$ 
Leveraging IK, we can then define a corresponding trajectory in configuration space as
$C = \langle\vec{q}_1, \vec{q}_2, ..., \vec{q}_t, ..., \vec{q}_{M}\rangle$ of $M$ waypoint configurations, assuming the trajectory is executed via linear interpolation in configuration space between the waypoint configurations.

We formulate our problem as a context learning problem.
Specifically, we consider the execution of tasks that can be defined as a desired motion of the robot's tip with respect to a context variable $\boldsymbol{\kappa}$, where generally $\boldsymbol{\kappa}$ is a vector of relevant scalar context values, e.g., the location of task-relevant objects and/or values identifying geometric properties of the robot's environment.
For this work, we consider $\boldsymbol{\kappa}$ as input given by the user.

We define a demonstration as a trajectory of reachable 3D robot tip positions paired with an instantiation of the context variable, for which the trajectory was gathered via human demonstration with known context.
The problem is then given a set of demonstrations as input as well as a new context, not before seen during the demonstrations, to output a configuration-space trajectory that performs the task, consistent with the demonstrations, under the new context.

%% file: approach.tex
We first collect demonstrations from a human that solve a specific task under varying contexts.
This produces a set of trajectory-context pairs, \(\{(T_i, \bm{\kappa}_i)\}_{i=1,\ldots,D}\) for a given task, and from which we can learn to generalize to new context.

We compare three models for learning from the demonstrations to solve the context-dependent task learning problem.
In the first, we leverage a linear ridge regression model to define a linear mapping between the context variables and a workspace trajectory, with weights learned from the demonstrations.
In the second, we replace the linear mapping with a non-linear mapping--utilizing a radial basis function kernel model.
The third model we present is a neural network to map the context vector to a workspace trajectory.
For all three models, once trained, we take the workspace trajectory predicted for the test-time context and use iterative inverse kinematics to produce a configuration-space trajectory that completes the task with the tendon-driven robot.

\subsection{Human Demonstration Collection}

In order to learn to autonomously execute the task, we collect a set of human demonstrations in the form of a sequence of robot tip positions.
Via a teleoperation setup, we provide a simulation environment in which a human moves the desired tip of the robot using a haptic input device, and the robot shape is interactively updated using IK (see Fig.~\ref{fig:high_level_figure}).

For each demonstration we vary the environment, corresponding to a change in the context variable $\boldsymbol{\kappa}$, and ask the user to demonstrate the task.
To do so, the human moves the tip of the simulated tendon-driven robot through the virtual environment.
We enable this by mapping the haptic-device tip position into the virtual environment and solving for a configuration that places the tip of the tendon-driven robot as close as possible to that position via IK.
We specifically leverage the FK model presented in~\cite{Rucker2011_TRO} to enable damped least squares iterative inverse kinematics~\cite{Wampler1986_TSMC}.
The user then records a sequence of robot tip positions that perform the desired task.
This produces one demonstration that pairs the environment, encoded via the context variable, to the trajectory.
We collect \(D\) such demonstrations each with different context variables $\boldsymbol{\kappa}_i$.

For the three models for learning, we choose to map the context to the demonstrations' tip \emph{positions}, rather than the configurations themselves.
We do so in order to reduce the complexity of the learning problem and to not require the methods to learn the tendon-driven robot's complex kinematics.
However, the tip positions in the collected demonstrations come from a simulated robot and its kinematics in order to ensure that the demonstrations contain only feasible robot tip positions.

\subsection{Learning to Map Context to a Trajectory}

We present three models that learn from the demonstrations and output a tip-space trajectory given a specific environmental context.

\paragraph*{Linear Ridge Regression}
For our first model, we utilize a linear ridge regression method~\cite{marquaridt1970generalized}.
We define a linear mapping between a context vector, \(\bm{\kappa}\), and an associated predicted tip-space trajectory $T^{\prime}$ via: $T^{\prime} = \boldsymbol{\kappa} W$,
where $\boldsymbol{\kappa}$ is the context variable and $W$ is a weight matrix to be learned. The dimension of $W$ is $k \times 3M$, where \(k\) is the size of $\boldsymbol{\kappa}$ and \(M\) is the number of waypoints in the trajectory.

In order to learn from the demonstrations, we optimize for the weights $W$ by solving:
\begin{equation}
\label{linear_loss}
\argmin_{W} \sum_{i=1}^{D}||\boldsymbol{\kappa}_{i}W - T_{i}||_{2}^{2} + \alpha||W||_{2}^{2},
\end{equation}
where $T_{i}$ is the i$^{th}$ demonstration trajectory, $\boldsymbol{\kappa_{i}}$ is the associated context, and \(D\) is the total number of demonstrations for the task.

\paragraph*{Kernel Ridge Regression}
We next present a model that replaces the linear mapping with a non-linear mapping via radial basis function (RBF) kernels~\cite{schaback2006kernel,beatson1999fast,poggio2002mathematical}.
Similar to the linear method, this method maps the context vector to a tip-space trajectory $T^{\prime}$, however we leverage a non-linear feature transform on top of the raw context variable, \(T^{\prime} = \phi(\boldsymbol{\kappa})W\).
Here again $\boldsymbol{\kappa}$ is the context variable and $W$ is a weight matrix to be learned.

We define the feature transform function, $\phi(\bm{\kappa})$ as the vector of kernel evaluations on the context variable \(\phi(\bm{\kappa}) = [k(\bm{\kappa}, x_{1}^{\prime}),\ldots,k(\bm{\kappa}, x_{D_k}^{\prime})]\), where $x_{j}^{\prime}$ is the j$^{th}$ kernel center and \(D_k\) is the total number of kernels.
We use the radial basis function kernel for all features
\begin{equation}
\label{kenerl_func}
k(\boldsymbol{\kappa}, x^{\prime}) = \exp(-\gamma ||\boldsymbol{\kappa} - x^{\prime}||^2).
\end{equation}
with the \(D_k\) kernel centers spaced throughout the context space. We learn the weights $W$ by solving the kernelized form of the ridge regression loss:
\begin{equation}
\label{rbf_loss}
\argmin_{W} \sum_{i=1}^{D}||\phi(\boldsymbol{\kappa}_{i})W - T_{i}||_{2}^{2} + \alpha||W||_{2}^{2},
\end{equation}
where $T_{i}$ is the i$^{th}$ demonstration trajectory, $\boldsymbol{\kappa}_{i}$ is the associated context, and D is the total number of demonstrations.

\paragraph*{Neural Trajectory Network}
Finally, we present a neural-network model to map the context to the tip-space trajectory.
Specifically, we utilize a feed forward neural network with a rectified linear unit~(ReLU) activation function that takes the context as input and outputs the tip-space trajectory as a vector of size $3M$ via $T^{\prime} = f(\boldsymbol{\kappa}, \theta)$, where $f$ is the neural network with learned parameters $\theta$.

We train the network on the demonstrations to optimize $\theta$ via a Mean Squared Error (MSE) loss function:
\begin{equation}
\label{traj_loss_comparison}
\argmin_{\theta} \frac{1}{3M\cdot D}{\sum_{i=1}^{D}||f(\boldsymbol{\kappa_{i}}, \theta) - T_{i}||_{2}^{2}},
\end{equation}
where $f(\boldsymbol{\kappa_{i}}, \theta)$ is the i$^{th}$ output from the trajectory network (with input $\boldsymbol{\kappa_{i}}$), $T_{i}$ is the i$^{th}$ demonstration trajectory and $D$ is the number of demonstrations each with $M$ waypoints.

\subsection{Task-Space Trajectory to Execution on the Robot}
Each of the three models outputs a task-space trajectory for a given context, however we must execute this trajectory on the robot.
To do so we leverage the IK function to produce the trajectory in configuration space, $C$, that closely follows the workspace trajectory $T^\prime$.

%% file: experiments.tex
\label{sec:experiments}

We demonstrate our method and evaluate its efficacy with three surgery-inspired tasks.
In the first, our method learns to trace a self-intersecting ``eight" shaped curve on the surface of a plane, where the widths and heights of the desired curves vary.
In the second, our method learns to trace a curve on the surface of two adjacent spheres with differing and varying radii, moving from the surface of one to the surface of the other.
These two tasks are inspired by the application of electrocautery or thermal ablation on the surface of curved or flat anatomy.
In the third, we place the robot in a simulated pleuroscopy scenario~\cite{Light2007_Book,Noppen2010_SRCCM} in a pleural volume segmented from a real patient.
In this environment our method learns to trace a specific curve on the surface of the pleural anatomy, where the anatomy's position and size relative to the robot varies.

\subsection{Learning to Trace a Curve on a Plane}\label{section_circle}

\begin{figure}[t]
    \centering
    \def\myPicHeight{0.35\columnwidth}
    \annotatedFigure[height=\myPicHeight]%
      {(a)}%
      {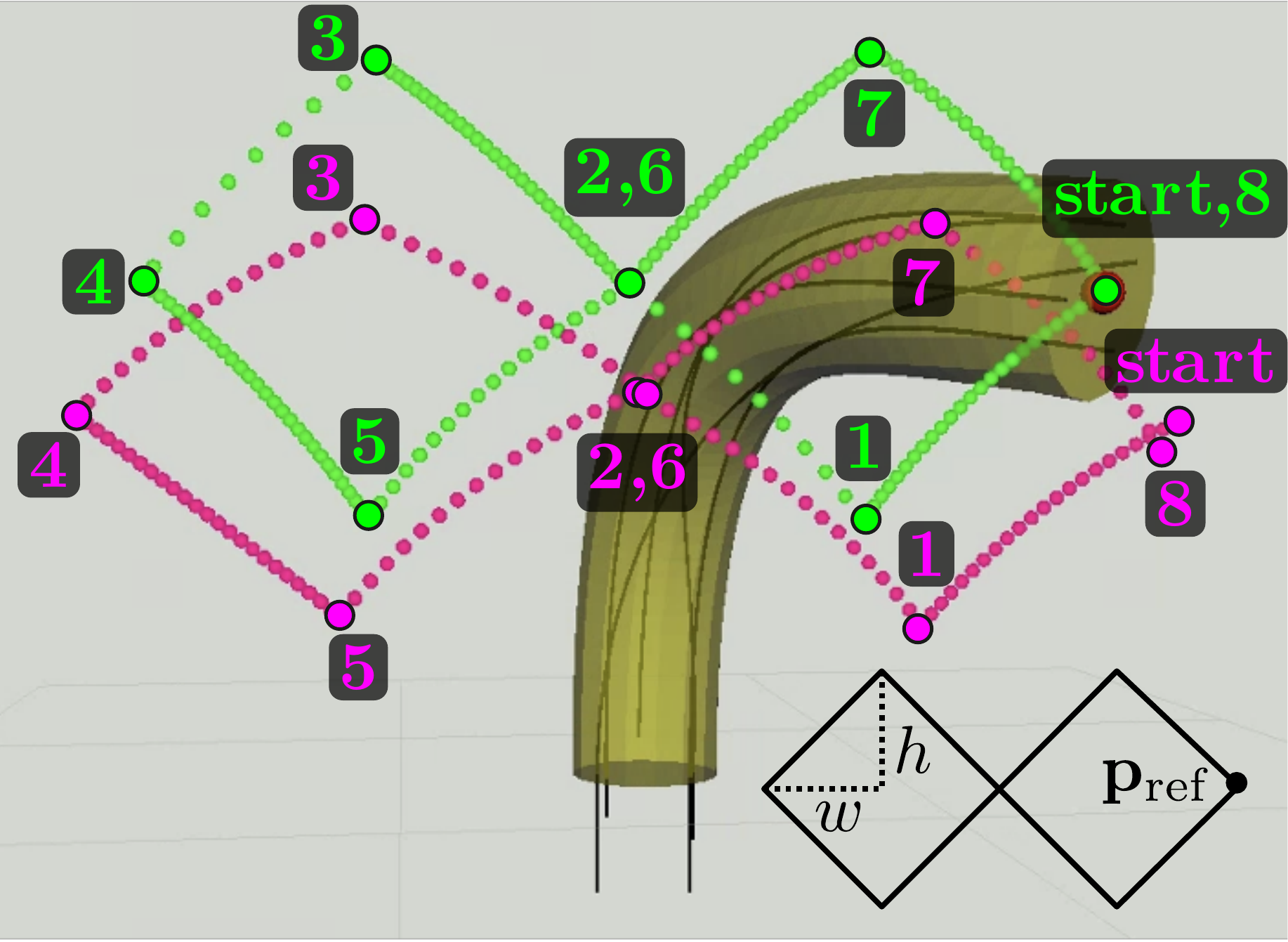}%
    \hfill %
    \annotatedFigure[height=\myPicHeight]%
      {(b)}%
      {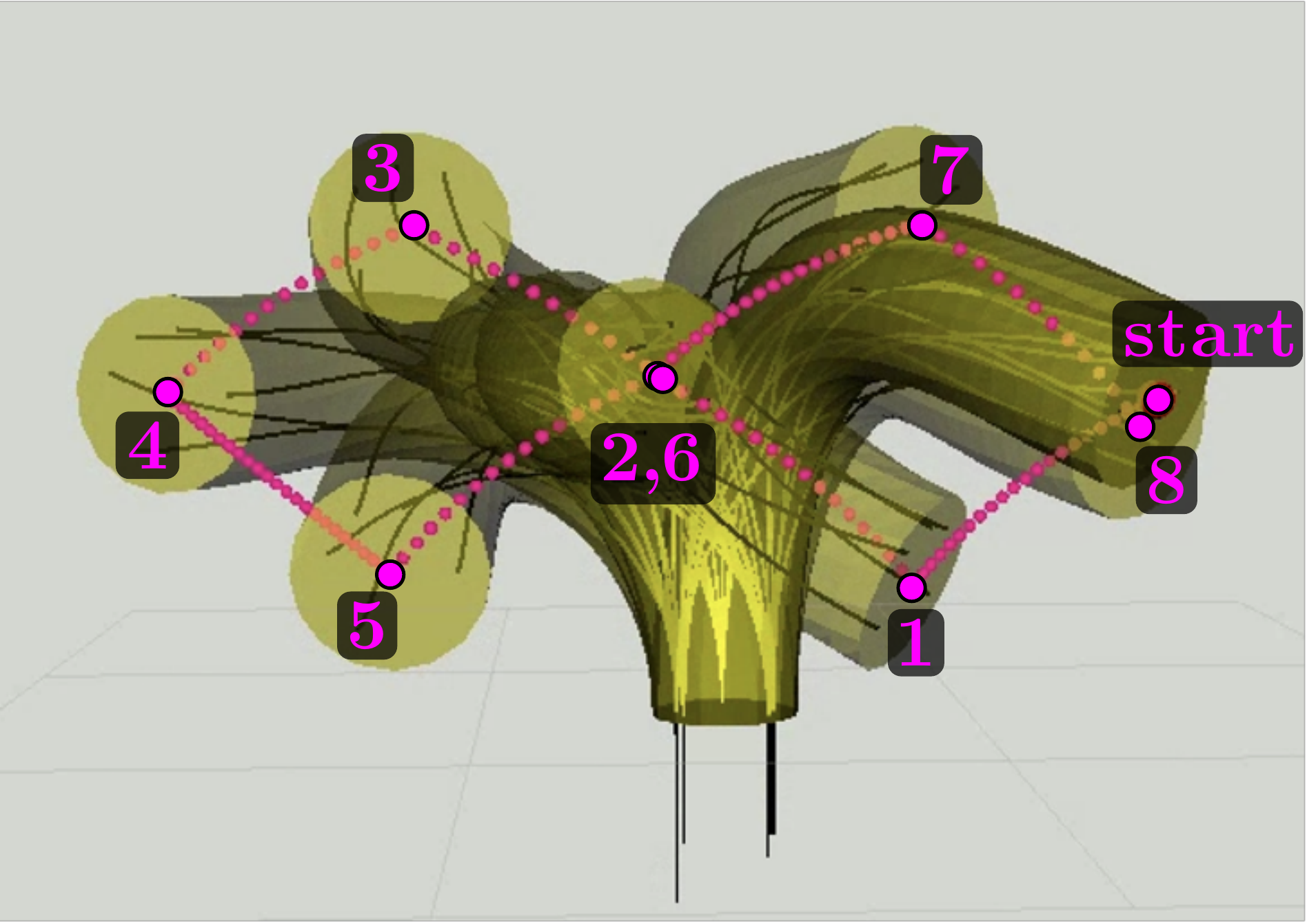}%
            \vspace{-0.5em}
      \caption{Visualization of the eight task. (a) The green trajectory represents a training trajectory gathered via demonstration and the pink trajectory represents a curve generated by the trajectory network approach given new context. (b) the robot configurations tracing the curve.}
      \vspace{-1.5em}
    \label{fig:eight_visualization_compare}
\end{figure}

In this task we evaluate the ability of our method to learn to trace a closed, ``eight" curve on a plane (see Fig.~\ref{fig:eight_visualization_compare}), inspired by the application of energy-based ablation.

We consider a robot of length $0.2$m with a $0.01$m radius.
The robot is routed with three straight tendons, distributed evenly around the backbone, and two helical tendons oppositely wrapped.
Both helically routed tendons make 0.64 revolutions from the base to the tip, routed in opposite directions.
Here we disable the robot's insertion/retraction and rotation degrees of freedom and leverage tendon tension only to control the robot.

For this task, we define the context as the starting point of the curve on the plane, defined as $\pref \in \mathbb{R}^3$, as well as the width $w$ and height $h$ parameters of the desired curve, such that $\boldsymbol{\kappa} = [\pref, w, h, 1]$ as shown in Fig.~\ref{fig:eight_visualization_compare}.
We augment the context with the scalar $1$ as we find it enables the learning of scalar bias for the models. 

To collect demonstrations with varying context, we sample uniformly at random 50 context variables with $\pref$ sampled from an $0.08$m by $0.04$m planar rectangle in the robot's workspace, $w$ from the range $(0.01$m$, 0.04$m$)$, and $h$  from $(0.01$m$, 0.04$m$)$.
For each sampled context variable, we collect a demonstration via the haptic device where a human moves the tip of the robot in the desired motion starting at $\pref$, with the scale of the curve defined by $w$ and $h$.

As the goal is to learn to trace the curve consistently with respect to the context variable, in order to evaluate performance we must first define a reference curve based on the demonstrations.
The ideal curve varies as the context varies, making it difficult to evaluate accuracy, however in this specific task we can express the curve as a function of the context variable via scaling.
Specifically, since we utilize width and height as aspects of the context variable, we can scale each demonstration to a single reference scale using $w$ and $h$. 
Given a specific trajectory $T = 
\langle p_1, p_2,\dots, p_{M}\rangle$, we compute a corresponding trajectory in the reference context $T_{\textrm{r}} = 
\langle p_1^{\prime}, p_2^{\prime},\dots, p_{M}^{\prime}\rangle$ via $p_{i}^{\prime} = \pref + (p_{i} - \pref)/([w \cdot 40,1,h \cdot 40])$.
This enables us to scale the demonstrations based on their specific context into a single reference context.
However, as the demonstrations were performed by a human, even in the reference context these curves deviate from each other to some extent.
We produce a single reference curve from these demonstrations by averaging the displacement of each waypoint on the curves in the reference context. 
This reference curve, along with a quantification of the variance in the demonstrations with respect to this curve will be used to compare and validate the output of our method---we wish for the method to exhibit similar variance to the demonstrations.

To evaluate our method's performance for each learning approach, we compare the robot's tip curve generated by the three versions of our method with this reference curve.
To do so, we utilize discrete Fr\'echet distance~\cite{Wylie2013_PhD}, a common metric for measuring the similarity between two 3-D space curves.
The goal then is for the method to produce robot motions that move the robot's tip, with respect to the context variable, in a way as similar to the reference curve as possible, ideally producing Fr\'echet distance values that are comparable to the distances between the individual demonstrations and the reference curve.

We first leverage this task to set the hyperparameters of the three learning approaches.
Under varying hyperparameters we train each model on $50$ demonstrations.
We then randomly sample 50 new context variables (from the same planar rectangle and the same $w$, $h$ ranges from which we sampled the demonstrations) and evaluate our method's performance in the scenarios defined by the new context.
For the linear ridge regression model, we choose from $\alpha \in \{0.01, 0.1, 1, 10\}$. 
We use the linear ridge regression implementation in scikit-learn~\cite{scikit-learn} and find via grid search the best performance when $\alpha = 0.01$.
For the RBF kernel model, we choose from $\alpha \in \{0.01, 0.1, 1, 10\}$ and $\gamma \in \{0.01, 0.1, 1, 10\}$. 
For this model we utilize the scikit-learn~\cite{scikit-learn} implementation of kernel ridge regression and find via grid search the best performance when $\gamma = 10$ and $\alpha = 0.01$.
The scikit-learn implementation utilizes $D_{k} = D$ kernel centers, i.e., the number of demonstrations, and each kernel center is set to one of the demonstrations' context variables, i.e., $x^{\prime}_{i} = \boldsymbol{\kappa}_{i}$.
For the trajectory network model, we investigate the effect of varying the neural network architecture, i.e., the number of hidden layers and the width of the hidden layers, on the method's performance.
We evaluate a range of architectures, as shown in Table~\ref{table:eight}, and find that a network with 2 hidden layers of 128 neurons each performs the best.
These parameters are used for all subsequent experiments.

\begin{table*}[t]
\centering
\caption{Average Fr\'echet distance to the reference trajectory for the eight task, under varying architectures (mean $\pm$ std in meters).}
\vspace{-1em}
\begin{tabular}{ccccccc}
\toprule
\multicolumn{7}{c}{Hidden layer architecture: number of hidden layers $\times$ width of hidden layers} \\
 \midrule
 2$\times$16 & 2$\times$32 & 2$\times$64 & 2$\times$128 & 3$\times$32 & 3$\times$64 & 3$\times$128\\
  0.0306 $\pm$ 0.0183 & 0.0288 $\pm$ 0.0211 & 0.0120 $\pm$ 0.0092 & \textbf{0.0098 $\pm$ 0.0065} &0.0134 $\pm$ 0.0068 & 0.0112 $\pm$ 0.00614 & 0.0137 $\pm$ 0.0069\\
 \bottomrule
\end{tabular}
\label{table:eight}
\vspace{-1em}
\end{table*}

\begin{figure}[t]
    \centering
    \includegraphics[width=\columnwidth]{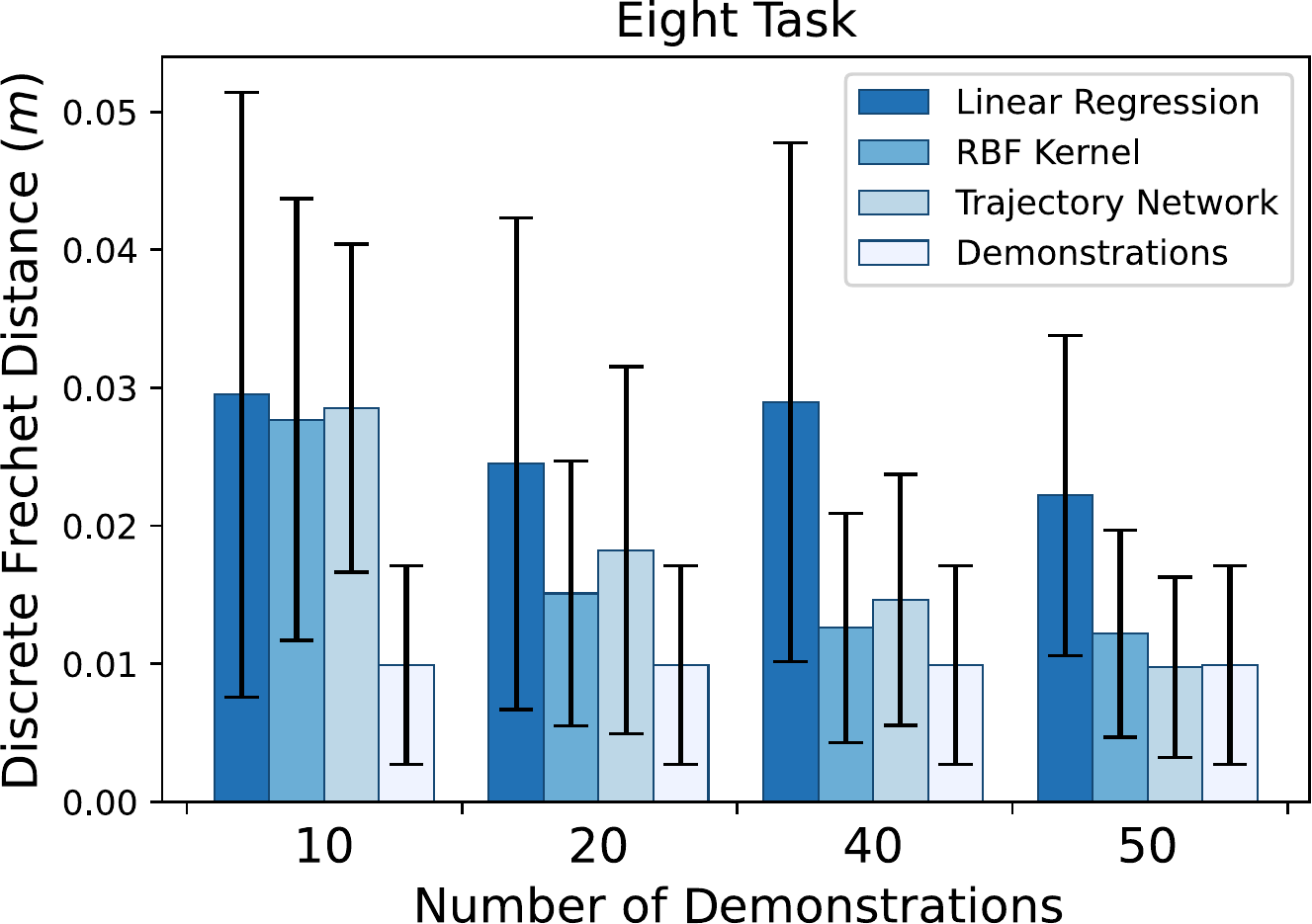}
    \vspace{-2em}
    \caption{Efficacy of our method using various learning approaches versus number of demonstrations.  Bars are the mean with error bars showing standard deviation.}
    \label{fig:eight_comparison}
    \vspace{-0.5em}
\end{figure}

We evaluate the performance of our method utilizing the three model classes as the number of demonstrations increases in Fig.~\ref{fig:eight_comparison}.
We train the three models on a subset of 50 demonstrations.
For each trained model, for each number of demonstrations, we randomly sample 50 context variables from the same ranges as sampled for the demonstrations and evaluate the performance of the approach via the discrete Fr\'echet distance compared to the reference trajectory.
As can be seen in Fig.~\ref{fig:eight_comparison}, all models generally improve with more training data, with the non-linear approaches outperforming the linear approach, and comparable performance between the RBF approach and the trajectory network.
Notably, the performance of our method using the trajectory network when trained on 50 demonstrations produces values with variance comparable to the variance exhibited by the human demonstrations themselves.

\subsection{Learning to Trace a Curve on the Surface of Spheres}
\label{section_sphere}

\begin{figure}[t]
    \centering
    \def\myPicHeight{0.48\columnwidth}
    \annotatedFigure[height=\myPicHeight]%
      {(a)}%
      {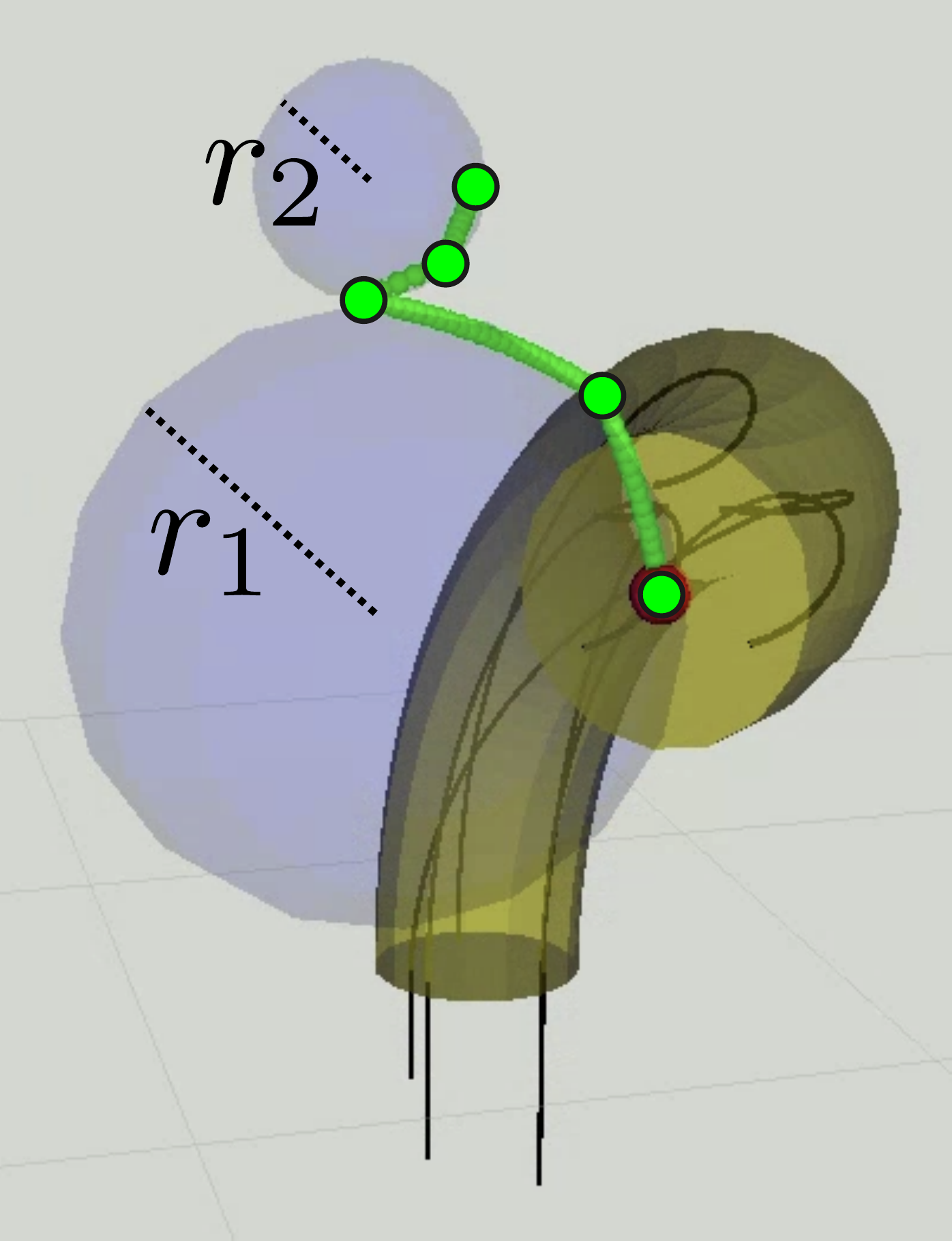}%
    \hfill%
    \annotatedFigure[height=\myPicHeight]%
      {(b)}%
      {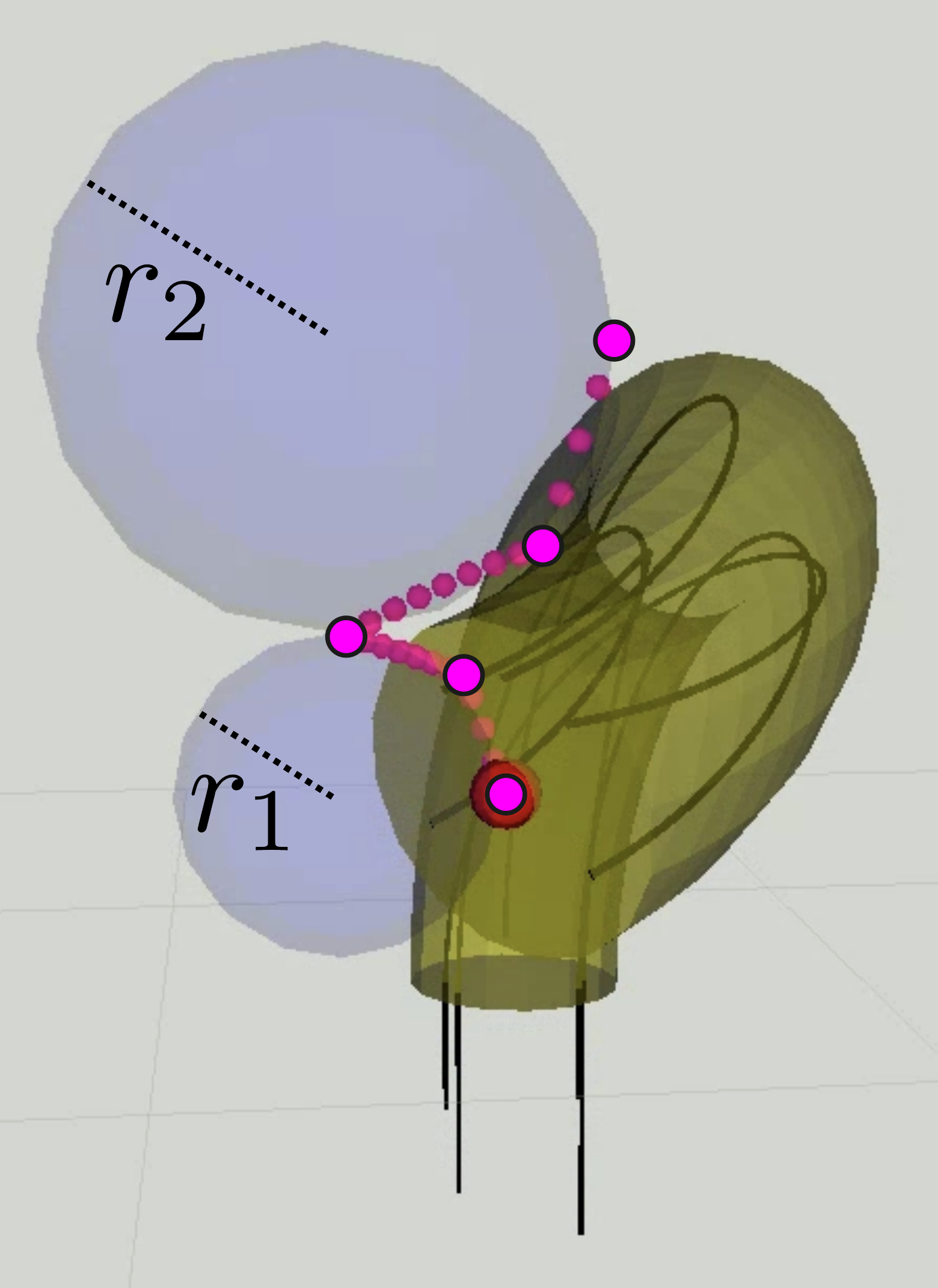}%
    \hfill%
    \annotatedFigure[height=\myPicHeight]%
      {(c)}%
      {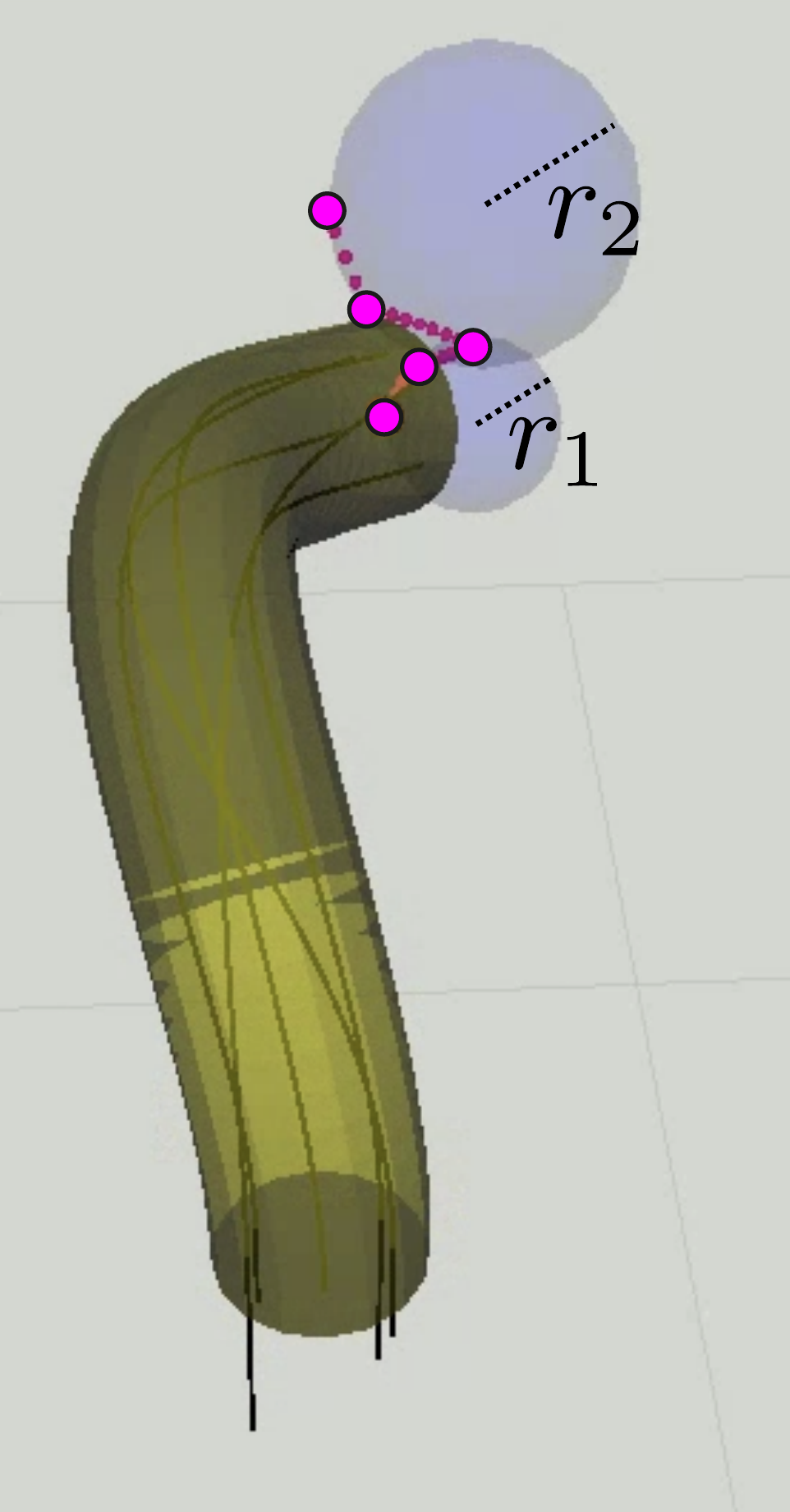}%
    \caption{Visualization of the double sphere task. (a) The curve (green) traced by a human teleoperating the robot--one of the training demonstrations. (b-c) Two views of the robot tracing a curve~(pink) generated by our method using the trajectory network approach under new context.}
    \label{fig:sphere_visualization_compare}
    \vspace{-2.0em}
\end{figure}

Similarly inspired by energy-based ablation, but on the surface of curved anatomy, in this task our method learns to trace a specific curve on the surface of two adjacent, vertically stacked spheres with differing radii (see Fig.~\ref{fig:sphere_visualization_compare}).

For this task we utilize the same robot described in Sec.~\ref{section_circle}.
We define the context as the start point of the curve, again denoted $\pref$, and the radii of the two spheres, $r_1$ and $r_2$, such that $\boldsymbol{\kappa} = [\pref, r_1, r_2, 1]$.

As the specifics of this task make it more difficult to generate a reference curve as in the previous task, here we evaluate the method's performance against specific human demonstrations.
We generate 20 training demonstrations and 10 demonstrations to be used for testing.
For each we randomly sample $\pref$ from a $0.04$m by $0.04$m plane in the robot's workspace, and separately sample the two radii from the range ($0.01$m, $0.03$m).
For each of the 30 randomly generated contexts we task a human to trace a curve on the sphere surfaces via teleoperating the robot, with the spheres visible in the user interface (see Fig.~\ref{fig:sphere_visualization_compare}).
We then train the three learning approaches on the 20 training demonstrations.

\begin{figure}[t]
    \centering
    \includegraphics[width=\columnwidth]{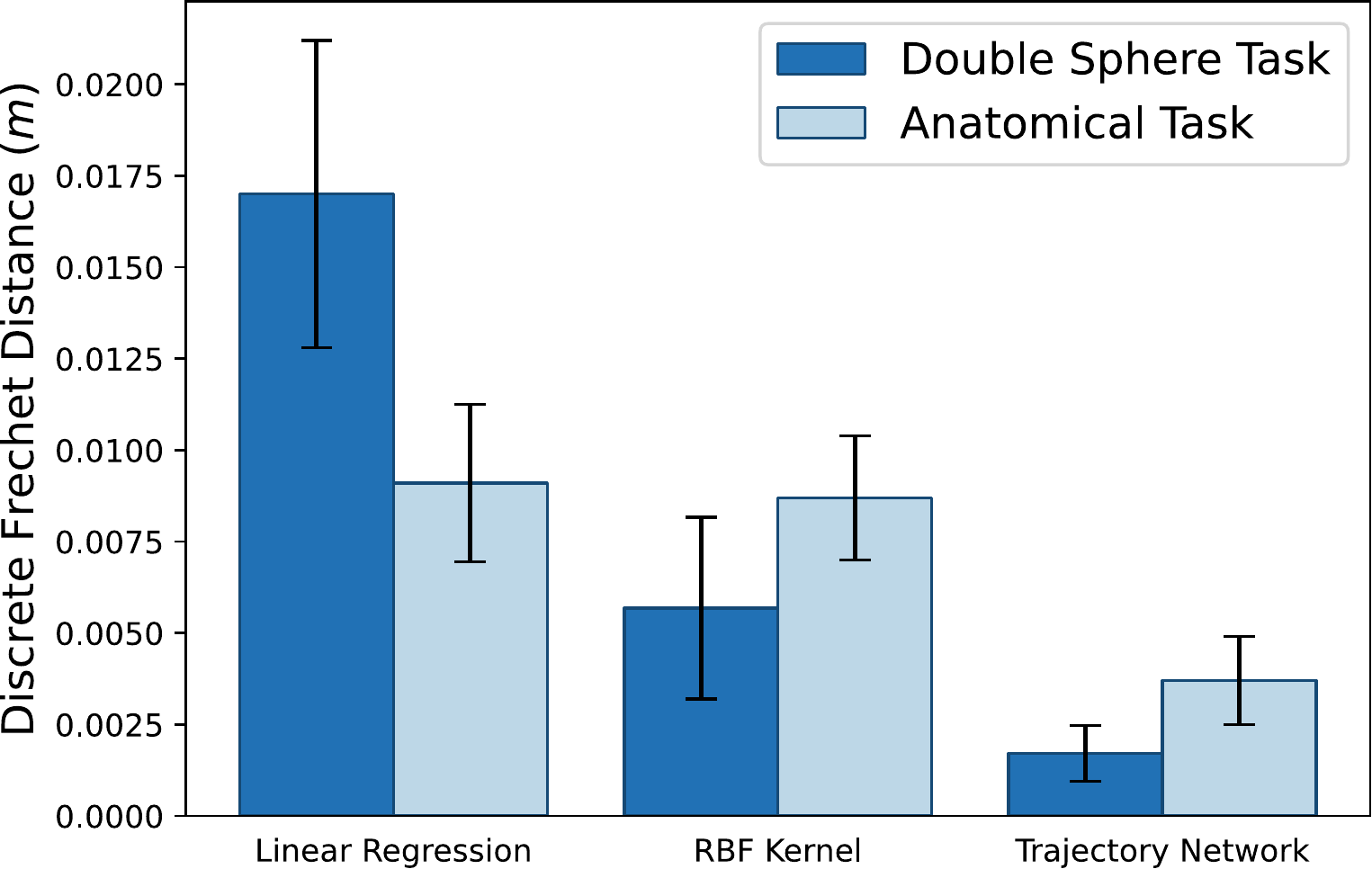}
    \vspace{-1.5em}
    \caption{A comparison of the different learning approaches on the double sphere task and the anatomical task.}
    \label{fig:ana_sphere_method_comparison}
\end{figure}

For each of the 10 test cases, we then task the learned models with generating target curves to be traced via IK and compare the traced curves with the human generated one.
Fig.~\ref{fig:sphere_visualization_compare} shows the task, an example of a training demonstration, and an example of the curve traced by the method utilizing the trajectory network approach.
We show the quantitative results for the three approaches on the 10 test cases in Fig.~\ref{fig:ana_sphere_method_comparison}.
Here we see that our method utilizing the kernel-based model outperforms the one using the linear approach, while the method utilizing the trajectory network model outperforms both.

\subsection{Learning to Trace the Surface of Anatomy}\label{section_ana}
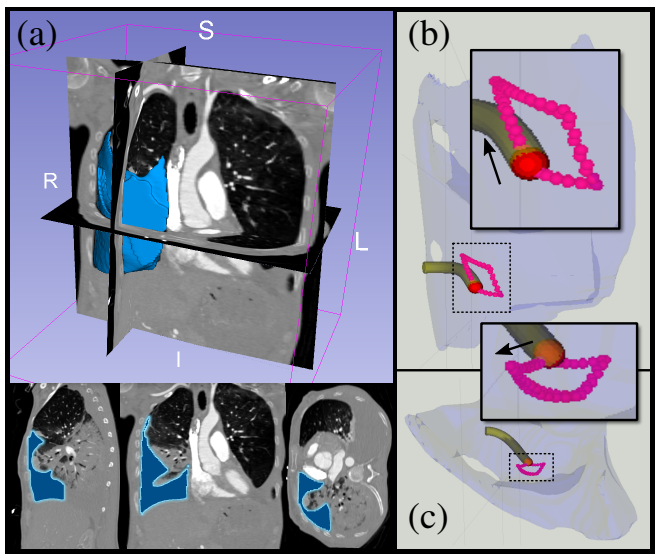
\begin{figure}[t]
    \centering
    \begin{tikzpicture}[above right, inner sep=0, outer sep=0]
      \node (image) at (0, 0)
        { \includegraphics[width=\columnwidth]{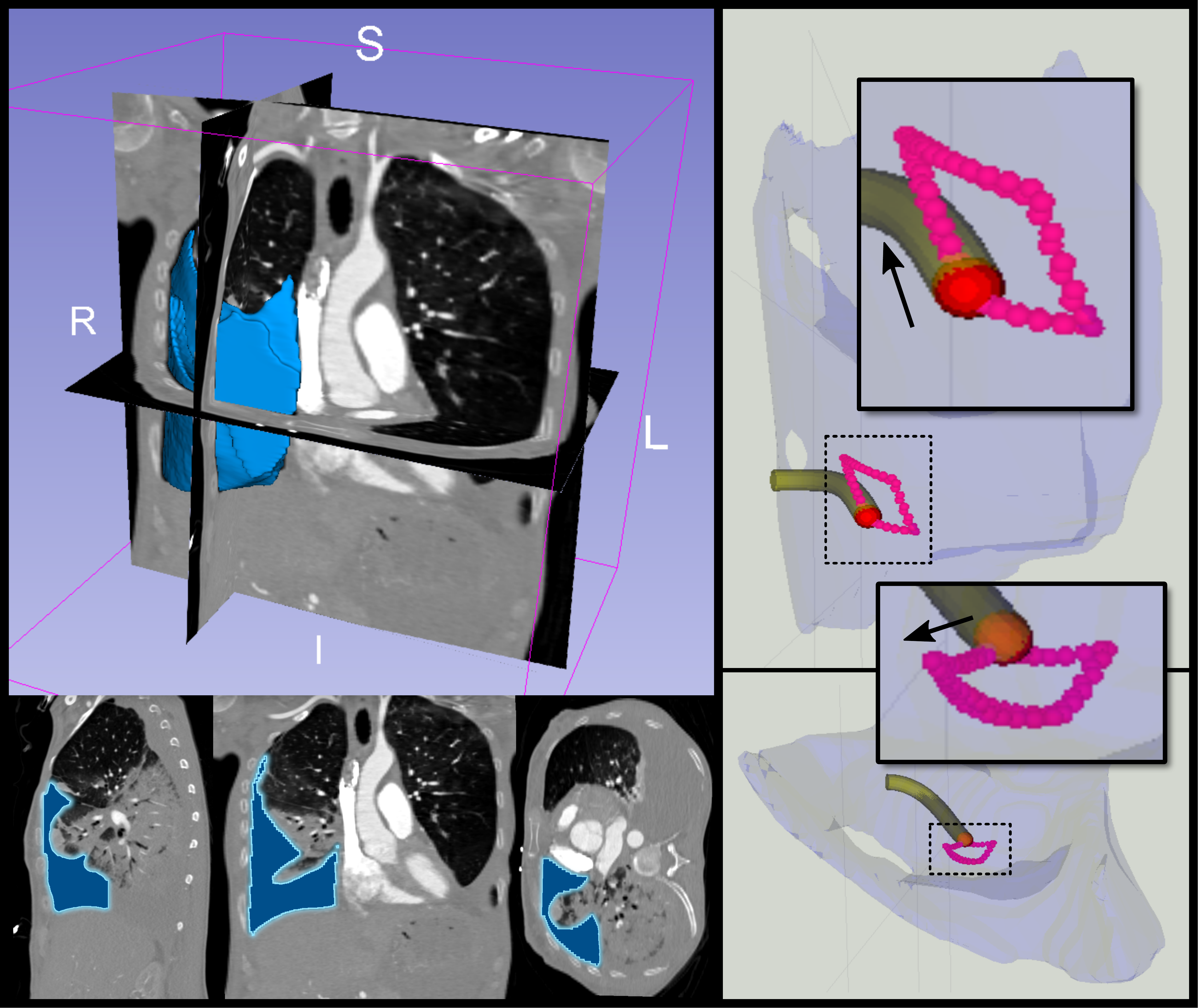} };
      \node[below right] (alabel) at (1.5mm, 71mm) {\Large(a)};
      \node[below right] (blabel) at (53.5mm, 71mm) {\Large(b)};
      \node[above right] (clabel) at (53.5mm, 2.5mm) {\Large(c)};
    \end{tikzpicture}
    \vspace{-1em}
    \caption{
      Anatomical environment scenario and task.
      (a) The anatomical environment was segmented from a pleural effusion CT scan via 3D Slicer~\cite{3DSlicer}.
      (b,c) Two different viewpoints of the trajectory network tracing the learned curve on the interior surface of the anatomical environment. The arrows show the start direction.
      }
      \vspace{-2em}
    \label{fig:ana_overview}
\end{figure}

Next we demonstrate a proof of concept of our method's ability to learn to trace the surface of patient anatomy in a simulated pleuroscopy task (see Fig.~\ref{fig:ana_overview}).
We use a simulation environment generated from a CT scan of a real patient undergoing this procedure, segmenting the boundaries of the air in the patient's pleural space using 3D Slicer~\cite{3DSlicer}.
We then simulate the tendon-driven robot operating inside the air volume in the patient's chest, between the chest wall and the collapsed lung.

In this task, we consider a slightly different tendon robot design, which is $0.20$m in length and $0.0025$m in radius.
We consider only one linearly routed tendon, with two oppositely routed helical tendons, 1.59 revolutions from base to tip.
We enable insertion/retraction as well as rotation for this task.

We define the context here as the anatomy's position relative to the robot's base, encoded via $\pref$, which is also the start point for the curve; as well as the scale of the anatomy relative to the robot's size, denoted as $s$, such that $\boldsymbol{\kappa} = [\pref, s, 1]$.
This choice of context is meant to simulate variation in the anatomy and in the robot's insertion pose into the pleural space.

We collect 20 training demonstrations and 10 test demonstrations wherein a human moves the robot in the pleural space to trace a diamond-shaped curve, first in one direction and then the reverse, on the interior surface of the anatomy (see Fig.~\ref{fig:ana_overview}).
For each of the demonstrations the context is randomly sampled as $s \sim \mathcal{U}(0.5,1.5)$, a multiplicative factor applied to the scale of the patient's segmented anatomy, and $\pref$ being perturbed from a nominal point with the perturbation sampled via $\sim \mathcal{U}(-0.01$m$,0.01$m$)$.
We then train the learning approaches on the 20 training demonstrations.

We then use each of our models to generate target curves given the context provided for the 10 test demonstrations. We compare the curve traced from the learned model with the human traced curves for those demonstrations.
The comparison results are shown in Fig.~\ref{fig:ana_sphere_method_comparison}.
Here we see that again the RBF kernel approach outperforms the linear approach, while the trajectory network approach outperforms both.
An example curve traced by the trajectory network approach is shown in pink in Fig.~\ref{fig:ana_overview}b and c.

%% file: Discussion.tex
In this work we take significant steps toward the automation of context-aware tendon-driven robot surgical tasks learned from human demonstrations.
We do so via a method leveraging learned models trained on demonstrations of the tasks where the context of the task was varied during the demonstrations.
The method is then able to perform the task successfully in situations unseen during training, e.g., when task relevant features are in a different location or given a different scale.
Our method performs best when utilizing a neural network-based model to generate target trajectories on two of the three surgery-inspired tasks, with the kernel-based model performing comparably in one of the tasks.
We also note that the trajectory network approach exhibits near-human-level performance in many cases.

In this work, we provide the method with the context variables as input.
However it is our intention, and a natural next step, to instead learn the context variables directly from the observed environment, e.g., learned from an endoscopic camera view or medical imaging.
We also intend to apply this concept to other continuum robots, such as concentric tube robots.
Along those lines, we intend to move further toward clinically applicable settings and tasks and utilize physical robots beyond simulation.

%% file: main.bbl
\begin{thebibliography}{10}
\providecommand{\url}[1]{#1}
\csname url@rmstyle\endcsname
\providecommand{\newblock}{\relax}
\providecommand{\bibinfo}[2]{#2}
\providecommand\BIBentrySTDinterwordspacing{\spaceskip=0pt\relax}
\providecommand\BIBentryALTinterwordstretchfactor{4}
\providecommand\BIBentryALTinterwordspacing{\spaceskip=\fontdimen2\font plus
\BIBentryALTinterwordstretchfactor\fontdimen3\font minus
  \fontdimen4\font\relax}
\providecommand\BIBforeignlanguage[2]{{%
\expandafter\ifx\csname l@#1\endcsname\relax
\typeout{** WARNING: IEEEtran.bst: No hyphenation pattern has been}%
\typeout{** loaded for the language `#1'. Using the pattern for}%
\typeout{** the default language instead.}%
\else
\language=\csname l@#1\endcsname
\fi
#2}}

\bibitem{Burgner2015_TRO}
J.~{Burgner-Kahrs}, D.~C. Rucker, and H.~Choset,
  ``\BIBforeignlanguage{en}{Continuum robots for medical applications: A
  survey},'' \emph{\BIBforeignlanguage{en}{IEEE Transactions on Robotics}},
  vol.~31, no.~6, pp. 1261--1280, Dec. 2015.

\bibitem{Webster2010_IJRR}
R.~J. Webster~III and B.~A. Jones, ``Design and kinematic modeling of constant
  curvature continuum robots: {{A}} review,'' \emph{The International Journal
  of Robotics Research}, vol.~29, no.~13, pp. 1661--1683, 2010.

\bibitem{Nguyen2015_IROS}
T.~Nguyen and J.~{Burgner-Kahrs}, ``A tendon-driven continuum robot with
  extensible sections,'' in \emph{{{IEEE}}/{{RSJ International Conference}} on
  {{Intelligent Robots}} and {{Systems}} ({{IROS}})}, Sept. 2015, pp.
  2130--2135.

\bibitem{Kato2015_TMECH}
T.~Kato, I.~Okumura, S.-E. Song, A.~J. Golby, and N.~Hata, ``Tendon-driven
  continuum robot for endoscopic surgery: Preclinical development and
  validation of a tension propagation model,'' \emph{IEEE/ASME Transactions on
  Mechatronics}, vol.~20, no.~5, pp. 2252--2263, Oct. 2015.

\bibitem{Kutzer2011_ICRA}
M.~D.~M. Kutzer, S.~M. Segreti, C.~Y. Brown, M.~Armand, R.~H. Taylor, and S.~C.
  Mears, ``Design of a new cable-driven manipulator with a large open lumen:
  {{Preliminary}} applications in the minimally-invasive removal of
  osteolysis,'' in \emph{Proc. {{IEEE}} Int. {{Conf}}. {{Robotics}} and
  Automation ({{ICRA}})}, May 2011, pp. 2913--2920.

\bibitem{Rucker2011_TRO}
D.~C. Rucker and R.~J. Webster~III, ``\BIBforeignlanguage{en}{Statics and
  dynamics of continuum robots with general tendon routing and external
  loading},'' \emph{\BIBforeignlanguage{en}{IEEE Transactions on Robotics}},
  vol.~27, no.~6, pp. 1033--1044, Dec. 2011.

\bibitem{Oliver-Butler2019_TRO}
K.~{Oliver-Butler}, J.~Till, and C.~Rucker, ``Continuum robot stiffness under
  external loads and prescribed tendon displacements,'' \emph{IEEE Transactions
  on Robotics}, vol.~35, no.~2, pp. 403--419, Apr. 2019.

\bibitem{yip2019robot}
M.~Yip and N.~Das, ``Robot autonomy for surgery,'' in \emph{The Encyclopedia of
  MEDICAL ROBOTICS: Volume 1 Minimally Invasive Surgical Robotics}.\hskip 1em
  plus 0.5em minus 0.4em\relax World Scientific, 2019, pp. 281--313.

\bibitem{argall2009survey}
B.~D. Argall, S.~Chernova, M.~Veloso, and B.~Browning, ``A survey of robot
  learning from demonstration,'' \emph{Robotics and Autonomous Systems},
  vol.~57, no.~5, pp. 469--483, 2009.

\bibitem{ravichandar2020recent}
H.~Ravichandar, A.~S. Polydoros, S.~Chernova, and A.~Billard, ``Recent advances
  in robot learning from demonstration,'' \emph{Annual Review of Control,
  Robotics, and Autonomous Systems}, vol.~3, pp. 297--330, 2020.

\bibitem{kober2011reinforcement}
J.~Kober, E.~Oztop, and J.~Peters, ``Reinforcement learning to adjust robot
  movements to new situations,'' \emph{Robotics: Science and Systems, MIT Press
  Journal}, vol.~6, pp. 33--40, 2011.

\bibitem{kumar2019contextual}
V.~Kumar, T.~Hermans, D.~Fox, S.~Birchfield, and J.~Tremblay, ``Contextual
  reinforcement learning of visuo-tactile multi-fingered grasping policies,''
  \emph{arXiv preprint arXiv:1911.09233}, 2019.

\bibitem{Michaud2010_Chest}
G.~Michaud, D.~M. Berkowitz, and A.~Ernst,
  ``\BIBforeignlanguage{en}{Pleuroscopy for diagnosis and therapy for pleural
  effusions},'' \emph{\BIBforeignlanguage{en}{Chest}}, vol. 138, no.~5, pp.
  1242--1246, Nov. 2010.

\bibitem{conkey2019learning}
A.~Conkey and T.~Hermans, ``Learning task constraints from demonstration for
  hybrid force/position control,'' in \emph{IEEE-RAS 19th International
  Conference on Humanoid Robots (Humanoids)}, 2019, pp. 162--169.

\bibitem{akgun2012keyframe}
B.~Akgun, M.~Cakmak, K.~Jiang, and A.~L. Thomaz, ``Keyframe-based learning from
  demonstration,'' \emph{International Journal of Social Robotics}, vol.~4,
  no.~4, pp. 343--355, 2012.

\bibitem{silver2013learning}
D.~Silver, J.~A. Bagnell, and A.~Stentz, ``Learning autonomous driving styles
  and maneuvers from expert demonstration,'' in \emph{Experimental
  Robotics}.\hskip 1em plus 0.5em minus 0.4em\relax Springer, 2013, pp.
  371--386.

\bibitem{kuderer2015learning}
M.~Kuderer, S.~Gulati, and W.~Burgard, ``Learning driving styles for autonomous
  vehicles from demonstration,'' in \emph{Proc. {{IEEE}} Int. {{Conf}}.
  {{Robotics}} and Automation ({{ICRA}})}, 2015, pp. 2641--2646.

\bibitem{mericcli2010biped}
{\c{C}}.~Meri{\c{c}}li and M.~Veloso, ``Biped walk learning through playback
  and corrective demonstration,'' in \emph{Proceedings of the AAAI Conference
  on Artificial Intelligence}, vol.~24, no.~1, 2010.

\bibitem{van2010superhuman}
J.~Van Den~Berg, S.~Miller, D.~Duckworth, H.~Hu, A.~Wan, X.-Y. Fu, K.~Goldberg,
  and P.~Abbeel, ``Superhuman performance of surgical tasks by robots using
  iterative learning from human-guided demonstrations,'' in \emph{Proc.
  {{IEEE}} Int. {{Conf}}. {{Robotics}} and Automation ({{ICRA}})}, 2010, pp.
  2074--2081.

\bibitem{murali2015learning}
A.~Murali, S.~Sen, B.~Kehoe, A.~Garg, S.~McFarland, S.~Patil, W.~D. Boyd,
  S.~Lim, P.~Abbeel, and K.~Goldberg, ``Learning by observation for surgical
  subtasks: Multilateral cutting of 3d viscoelastic and 2d orthotropic tissue
  phantoms,'' in \emph{Proc. {{IEEE}} Int. {{Conf}}. {{Robotics}} and
  Automation ({{ICRA}})}, 2015, pp. 1202--1209.

\bibitem{kim2020autonomously}
J.~W. Kim, C.~He, M.~Urias, P.~Gehlbach, G.~D. Hager, I.~Iordachita, and
  M.~Kobilarov, ``Autonomously navigating a surgical tool inside the eye by
  learning from demonstration,'' in \emph{Proc. {{IEEE}} Int. {{Conf}}.
  {{Robotics}} and Automation ({{ICRA}})}, 2020, pp. 7351--7357.

\bibitem{fong2018kinesthetic}
J.~Fong and M.~Tavakoli, ``Kinesthetic teaching of a therapist's behavior to a
  rehabilitation robot,'' in \emph{International Symposium on Medical Robotics
  (ISMR)}.\hskip 1em plus 0.5em minus 0.4em\relax IEEE, 2018, pp. 1--6.

\bibitem{najafi2017robotic}
M.~Najafi, M.~Sharifi, K.~Adams, and M.~Tavakoli, ``Robotic assistance for
  children with cerebral palsy based on learning from tele-cooperative
  demonstration,'' \emph{International Journal of Intelligent Robotics and
  Applications}, vol.~1, no.~1, pp. 43--54, 2017.

\bibitem{xu2017data}
W.~Xu, J.~Chen, H.~Y. Lau, and H.~Ren, ``Data-driven methods towards learning
  the highly nonlinear inverse kinematics of tendon-driven surgical
  manipulators,'' \emph{The International Journal of Medical Robotics and
  Computer Assisted Surgery}, vol.~13, no.~3, p. e1774, 2017.

\bibitem{giorelli2015neural}
M.~Giorelli, F.~Renda, M.~Calisti, A.~Arienti, G.~Ferri, and C.~Laschi,
  ``Neural network and jacobian method for solving the inverse statics of a
  cable-driven soft arm with nonconstant curvature,'' \emph{IEEE Transactions
  on Robotics}, vol.~31, no.~4, pp. 823--834, 2015.

\bibitem{Bergeles2015_Hamlyn}
C.~Bergeles, F.~Y. Lin, and G.~Z. Yang, ``Concentric tube robot kinematics
  using neural networks,'' in \emph{Hamlyn Symposium on Medical Robotics}, June
  2015, pp. 1--2.

\bibitem{Grassmann2018_IROS}
R.~Grassmann, V.~Modes, and J.~{Burgner-Kahrs}, ``Learning the forward and
  inverse kinematics of a 6-{{DOF}} concentric tube continuum robot in
  {{SE}}(3),'' in \emph{{{IEEE}}/{{RSJ}} International Conference on
  Intelligent Robots and Systems ({{IROS}})}, {Madrid, Spain}, Oct. 2018, pp.
  5125--5132.

\bibitem{kuntz2020learning}
A.~Kuntz, A.~Sethi, R.~J. Webster, and R.~Alterovitz, ``Learning the complete
  shape of concentric tube robots,'' \emph{IEEE Transactions on Medical
  Robotics and Bionics}, vol.~2, no.~2, pp. 140--147, 2020.

\bibitem{Donat2020_TMRB}
H.~Donat, S.~Lilge, J.~{Burgner-Kahrs}, and J.~J. Steil, ``Estimating tip
  contact forces for concentric tube continuum robots based on backbone
  deflection,'' \emph{IEEE Transactions on Medical Robotics and Bionics},
  vol.~2, no.~4, pp. 619--630, Nov. 2020.

\bibitem{iyengar2020investigating}
K.~Iyengar, G.~Dwyer, and D.~Stoyanov, ``Investigating exploration for deep
  reinforcement learning of concentric tube robot control,''
  \emph{International Journal of Computer Assisted Radiology and Surgery},
  vol.~15, pp. 1157--1165, 2020.

\bibitem{Wampler1986_TSMC}
C.~W. Wampler, ``Manipulator inverse kinematic solutions based on vector
  formulations and damped least-squares methods,'' \emph{IEEE Trans. Systems,
  Man and Cybernetics}, vol.~16, no.~1, pp. 93--101, 1986.

\bibitem{marquaridt1970generalized}
D.~W. Marquaridt, ``Generalized inverses, ridge regression, biased linear
  estimation, and nonlinear estimation,'' \emph{Technometrics}, vol.~12, no.~3,
  pp. 591--612, 1970.

\bibitem{schaback2006kernel}
R.~Schaback and H.~Wendland, ``Kernel techniques: from machine learning to
  meshless methods,'' \emph{Acta numerica}, vol.~15, p. 543, 2006.

\bibitem{beatson1999fast}
R.~K. Beatson, J.~B. Cherrie, and C.~T. Mouat, ``Fast fitting of radial basis
  functions: Methods based on preconditioned gmres iteration,'' \emph{Advances
  in Computational Mathematics}, vol.~11, no.~2, pp. 253--270, 1999.

\bibitem{poggio2002mathematical}
T.~Poggio and C.~R. Shelton, ``On the mathematical foundations of learning,''
  \emph{American Mathematical Society}, vol.~39, no.~1, pp. 1--49, 2002.

\bibitem{Light2007_Book}
R.~W. Light, \emph{Pleural Diseases}.\hskip 1em plus 0.5em minus 0.4em\relax
  {Lippincott Williams \& Wilkins}, 2007.

\bibitem{Noppen2010_SRCCM}
M.~Noppen, ``The utility of thoracoscopy in the diagnosis and management of
  pleural disease,'' in \emph{Seminars in Respiratory and Critical Care
  Medicine}, vol.~31, 2010, pp. 751--759.

\bibitem{Wylie2013_PhD}
T.~R. Wylie, ``The discrete {{Fr\'echet}} distance with applications,'' {{PhD
  Thesis}}, Montana State University-Bozeman, College of Engineering, 2013.

\bibitem{scikit-learn}
F.~Pedregosa, G.~Varoquaux, A.~Gramfort, V.~Michel, B.~Thirion, O.~Grisel,
  M.~Blondel, P.~Prettenhofer, R.~Weiss, V.~Dubourg, J.~Vanderplas, A.~Passos,
  D.~Cournapeau, M.~Brucher, M.~Perrot, and E.~Duchesnay, ``Scikit-learn:
  machine learning in {P}ython,'' \emph{Journal of Machine Learning Research},
  vol.~12, pp. 2825--2830, 2011.

\bibitem{3DSlicer}
A.~Fedorov, R.~Beichel, J.~{Kalpathy-Cramer}, J.~Finet, J.-C. {Fillion-Robin},
  S.~Pujol, C.~Bauer, D.~Jennings, F.~Fennessy, M.~Sonka, J.~Buatti,
  S.~Aylward, J.~V. Miller, S.~Pieper, and R.~Kikinis, ``{{3D Slicer}} as an
  image computing platform for the {{Quantitative Imaging Network}}.''
  \emph{Magnetic Resonance Imaging}, vol.~30, no.~9, pp. 1323--1341, 2012.

\end{thebibliography}
